\documentclass[preprint,12pt]{elsarticle}

\usepackage{amssymb}
\usepackage{amsmath}
\usepackage{booktabs}
\usepackage{multirow}
\usepackage{amsthm}

\journal{}

\begin{document}

\begin{frontmatter}


\author[1]{Jiahuan Long}
\author[1]{Wenzhe Zhang}
\author[1]{Ning Wang}
\author[1]{Tingsong Jiang}
\author[1]{Wen Yao}

\affiliation[1]{
  organization={Chinese Academy of Military Sciences},
  city={Beijing},
  country={China}
}

\title{FR-Mamba: Time-Series Physical Field Reconstruction Based on State Space Model}




\begin{abstract}
Physical field reconstruction (PFR) aims to predict the state distribution of physical quantities (e.g., velocity, pressure, and temperature) based on limited sensor measurements. It plays a critical role in domains such as fluid dynamics and thermodynamics.  However, existing deep learning methods often fail to capture long-range temporal dependencies, resulting in suboptimal performance on time-evolving physical systems. To address this, we propose FR-Mamba, a novel spatiotemporal flow field reconstruction framework based on state space modeling. Specifically, we design a hybrid neural network architecture that combines Fourier Neural Operator (FNO) and State Space Model (SSM) to capture both global spatial features and long-range temporal dependencies.
We adopt Mamba, a recently proposed efficient SSM architecture, to model long-range temporal dependencies with linear time complexity.
In parallel, the FNO is employed to capture non-local spatial features by leveraging frequency-domain transformations.
The spatiotemporal representations extracted by these two components are then fused to reconstruct the full-field distribution of the physical system.
Extensive experiments demonstrate that our approach significantly outperforms existing PFR methods in flow field reconstruction tasks, achieving high-accuracy performance on long sequences.

\end{abstract}




\end{frontmatter}



\section{Introduction}
\label{sec1}
Flow field reconstruction from sparse sensor measurements is a fundamental yet challenging problem in fluid mechanics and engineering applications~\cite{yu2019flowfield, shu2023physics, deng2022dual, luo2023reconstruction}. 
High-resolution reconstruction of unsteady flow fields is essential for understanding the underlying fluid dynamics and supporting complex decision-making in real-world systems. A classic example is the two-dimensional flow past a cylinder~\cite{nguyen2023state}, which exhibits periodic vortex shedding and is commonly used to model real-world flows around structures like buildings or bridge columns. 
However, due to the sparse and uneven placement of sensors, traditional statistical methods~\cite{das2012simulated, zhou2015compressed, loiseau2018sparse} often fail to capture fine-grained structures—particularly in regions with rich dynamics.

In recent years, deep learning-based physical reconstruction models have made progress in mapping sensor inputs to global flow field outputs. However, most existing methods operate in a frame-by-frame manner, relying solely on single-time-step observations and thus ignoring the rich temporal context embedded in historical measurements. 
To address this, some approaches have introduced sequence models such as LSTM~\cite{LSTM}, RNN~\cite{sherstinsky2020fundamentals} and Transformer~\cite{PFR-Transformer} to capture temporal dependencies. While these models offer a step forward, they often struggle with memory degradation and vanishing gradients when dealing with long sequences, limiting their ability to model long-range temporal dynamics effectively. Therefore, it is important to design more efficient and scalable network architectures capable of modeling long-range temporal dependencies in physical fields.

State space models (SSMs)~\cite{ssmS4, ssmhypothesis, mamba, mambavision, li2023nlpmamba, umatani2023time} have recently gained attention for their ability to efficiently model long-range dependencies with linear time and memory complexity. Among them, Mamba~\cite{mamba} has emerged as one of the most powerful and widely adopted SSM architectures to date. It is designed for efficient long-sequence modeling, combining the expressive power of state-space models with a streamlined architecture that enables fast and scalable training in practice. For instance, Vision-Mamba improves video sequence modeling via efficient state-space representations~\cite{gu2023visionmamba}, while NLP-Mamba achieves state-of-the-art results in long-text processing~\cite{li2023nlpmamba}. These advancements motivate us to explore whether SSMs can be effectively adapted to the domain of physical field reconstruction.

However, directly applying existing SSMs to time-series physical field reconstruction has inherent limitations. In natural language processing, each token carries semantic meaning, and SSMs primarily model the temporal evolution of these meanings. In contrast, physical sensor data is inherently spatial, with each measurement tied to a fixed spatial location.  Flattening such data into a 1D sequence disrupts the spatial structure, making it difficult for pure SSMs to capture inter-sensor spatial correlations—thus limiting reconstruction performance.

To address the spatial modeling limitations, recent studies have explored frequency-domain approaches such as Fourier Neural Operators (FNOs)~\cite{li2020fourier}, which offer strong generalization and expressive power for spatial field reconstruction. Traditionally, FNOs have been combined with Transformer- or RNN-based temporal modules~\cite{pathak2022fourcastnet, wang2023predrnn}. However, these architectures often suffer from high computational cost and limited scalability in long-sequence scenarios due to quadratic attention complexity or recurrent processing constraints. In contrast, modern SSMs like Mamba provide a linear-time, memory-efficient alternative for capturing long-range temporal dependencies. Motivated by this complementarity, we integrate FNO and SSM into a unified framework that enables efficient and expressive modeling of both spatial and temporal structures in physical fields.

In this paper, we propose FR-Mamba, a Mamba-based model for time-series physical field reconstruction. Specifically, we design an enhanced architecture that integrates a Fourier Neural Operator (FNO) with a State Space Model (SSM). The SSM branch efficiently captures long-range temporal dependencies from sequential sensor data, while the FNO branch models global spatial structures through frequency-domain representations. This dual-branch design enables the model to recover implicit spatial correlations even when inputs are flattened into 1D sequences. By stacking multiple Mamba blocks, FR-Mamba attains strong representational capacity to address complex, large-scale flow reconstruction tasks. Extensive experiments demonstrate that FR-Mamba significantly outperforms existing methods, particularly in long-sequence prediction settings. The main contributions are as follows:


\begin{itemize}
    \item To the best of our knowledge, it is the first work to introduce the State Space Model (SSM) into physical field reconstruction. Leveraging SSM’s long-sequence modeling ability, we can capture historical dependencies from sensor data with linear time complexity.

    \item We propose a novel dual-branch FNO-Mamba architecture that integrates Fourier Neural Operators (FNOs) with Mamba to jointly learn spatial and temporal features. The Mamba branch encodes long-range temporal dynamics, while the FNO branch captures global spatial structures through frequency-domain representations.

    \item  Extensive experiments on flow field datasets demonstrate that our method significantly outperforms existing approaches in terms of global reconstruction error. It maintains stable accuracy over long sequences and generalizes well across diverse flow regimes.
\end{itemize}

\section{Related Work}

\subsection{Deep Learning Approaches for 
Physical Field Reconstruction}
Physical field reconstruction aims to estimate the spatiotemporal distribution of physical quantities (e.g., velocity, pressure, temperature) based on sparse sensor measurements, the applications of which include geophysics~\cite{huang2023deep, tian2005multiple}, fluid dynamics~\cite{VoronoiCNN, loiseau2018sparse}, thermodynamics~\cite{peng2022deep, chen2023machine} and climate science~\cite{su2022subsurface}.
Traditional approaches to field reconstruction often rely on physics-based partial differential equations (PDEs)~\cite{gherlone2012shape, gu2017application} and statistical methods~\cite{das2012simulated, zhou2015compressed, loiseau2018sparse}. While these methods can model some phenomena, they face challenges in integrating sparse sensor data, especially when dealing with complex or nonlinear systems. 
In recent years, deep learning-based models have become mainstream for physical field reconstruction~\cite{GappyMLP, Voronoi-UNet, VoronoiCNN, shallowdecoder, PFR-Transformer}. GappyMLP~\cite{GappyMLP} proposes a hybrid approach that combines POD with multilayer perceptrons to reconstruct physical fields from sparse sensor measurements. VoronoiUNet~\cite{Voronoi-UNet} introduces a Voronoi-encoded Unet architecture with uncertainty quantification for reliable temperature field prediction under irregular sensor layouts. VoronoiCNN~\cite{VoronoiCNN} utilizes Voronoi tessellation to convert unstructured sensor inputs into grid-based representations, enabling CNN-based spatial flow reconstruction. ShallowDecoder~\cite{shallowdecoder} employs a shallow feedforward neural network to directly map sparse observations to high-dimensional flow fields with minimal model complexity. PFTransformer~\cite{PFR-Transformer} employs an attention-based encoder–decoder framework to reconstruct physical fields from sparse and irregular sensor inputs, offering scalable and efficient inference for large spatiotemporal domains.
However, these PFR methods suffer from limitations such as vanishing gradients and a failure to effectively capture long-range dependencies over extended sequences. These challenges make them less suitable for reconstructing time-series physical fields, where capturing long-term temporal correlations is essential.

\subsection{State Space Models and Mamba}
State-space models (SSMs)~\cite{ssmS4, ssmhypothesis, mamba, mambavision, li2023nlpmamba} have gained increasing attention in the machine learning community for their ability to efficiently model long-term temporal dependencies. In time-series modeling, SSMs provide a framework for compressing and tracking temporal information, offering significant advantages over traditional sequence models such as LSTM and RNN. Early works like LSSL~\cite{LSSL} and S4~\cite{ssmS4} introduced efficient state representations and linear scalability with sequence length, overcoming limitations of CNNs and transformers. Later advancements such as S5~\cite{s5} and H3~\cite{h3} improved computational efficiency and achieved competitive performance in tasks like language modeling.
Among the recent advances, Mamba~\cite{mamba} stands out as a state-of-the-art SSM architecture that achieves strong performance across various domains, including video processing and natural language understanding. By introducing selective state-space representations, Mamba enables efficient long-sequence processing while preserving modeling capacity. For instance, Vision-Mamba~\cite{gu2023visionmamba} improves real-time video processing by compressing long video sequences into compact representations. Similarly, NLP-Mamba~\cite{li2023nlpmamba} enhances long-text understanding and generation by capturing dependencies across sentences and paragraphs. In this paper, we explore the application of state-space models to time-series physical field reconstruction, aiming to capture long-range temporal dependencies from sparse sensor measurements in complex dynamical systems.

\section{Methodology}
In this section, we first introduce the problem formulation and define the training objective for reconstructing vorticity fields from sparse sensor observations (Section~3.1). We then outline the overall FR-Mamba architecture, emphasizing its hierarchical structure and dual-branch design for temporal and spatial feature learning (Section~3.2). Section~3.3 details the FNO-Mamba module, which combines selective state-space modeling with frequency-domain spatial encoding. Finally, Section~3.4 presents the complete flow reconstruction pipeline, including spatiotemporal feature fusion and the final decoding process for generating dense physical field outputs.

\subsection{Problem Formulation}

The goal of this work is to reconstruct time-series vorticity fields from sparse sensor observations. Vorticity is a key physical quantity in fluid dynamics that characterizes local rotational motion within a flow field. Accurately recovering vorticity over time enables better understanding and modeling of transient flow behavior.

At each time step \(t \in \{1, \dots, T\}\), we are given a sparse set of vorticity measurements \(x^t \in \mathbb{R}^{N_s}\), where \(N_s\) is the number of sensors. The target is to reconstruct the full vorticity field \(u^t \in \mathbb{R}^{H \times W}\), representing the underlying flow dynamics on a 2D spatial grid. A neural network model \(f_\theta\) is learned to map the input sequence \(\{x^1, \dots, x^T\}\) to the corresponding predictions \(\{\hat{u}^1, \dots, \hat{u}^T\}\). The model is trained by minimizing the mean absolute error (MAE) between predicted and ground-truth fields, defined as:

\[
\mathcal{L}_{\text{MAE}} = \frac{1}{THW} \sum_{t=1}^T \sum_{i=1}^{H} \sum_{j=1}^{W} \left| \hat{u}_{i,j}^t - u_{i,j}^t \right|.
\]

\begin{figure}[htbp]
	\centering
	\includegraphics[width=0.95\textwidth]{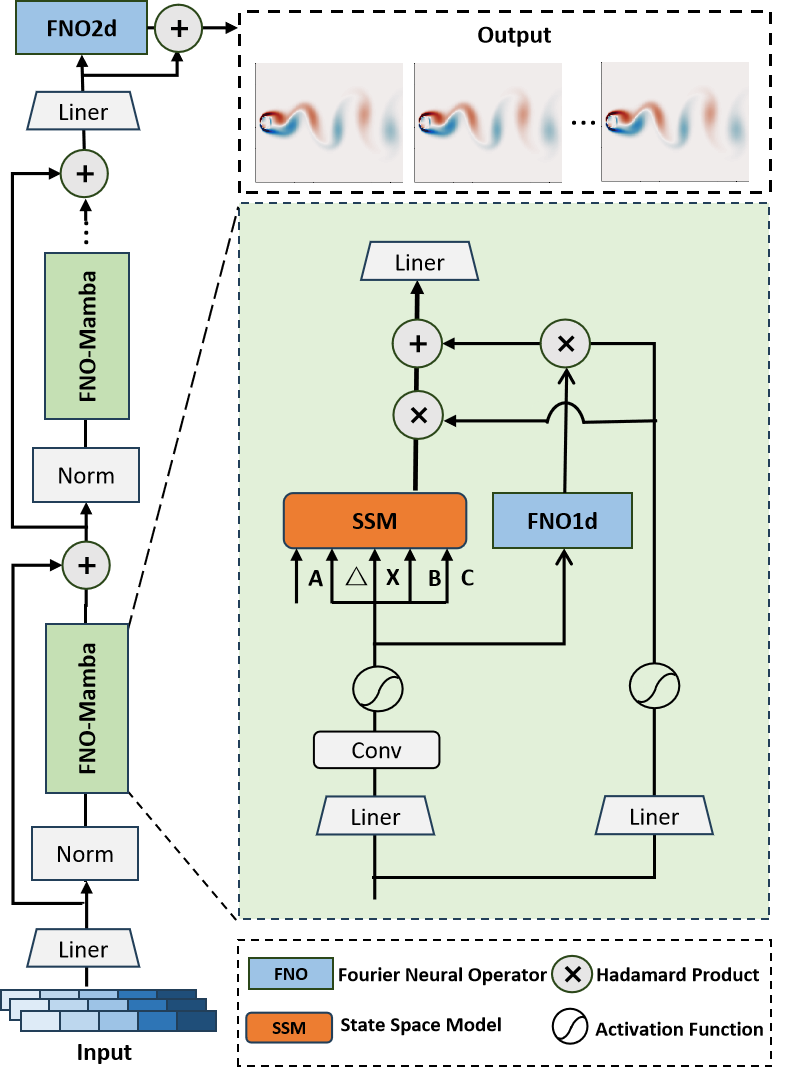}  
	\caption{The framework of our FR-Mamba.}
	\label{fig：framework}
\end{figure}

\subsection{Overview of Our Framework}
As illustrated in Figure~\ref{fig：framework}, we present the overall architecture of FR-Mamba for reconstructing time-series physical fields from sparse observations. The model is composed of stacked FNO-Mamba blocks that support hierarchical representation learning.
Each FNO-Mamba block comprises two branches dedicated to temporal and spatial feature extraction. Specifically, the temporal feature branch employs a State Space Model (SSM) to efficiently extract historical temporal features and state representations from sensor input sequences. The spatial feature branch utilizes a one-dimensional Fourier Neural Operator (FNO1d) to transform sensor data into the frequency domain, effectively capturing global spatial features. Finally, the high-dimensional temporal and spatial features extracted by the two branches within each FNO-Mamba block are fused through a gating mechanism. The fused spatiotemporal representation is then mapped into the target transient flow field using a feedforward neural network. To further enhance global spatial consistency, a two-dimensional Fourier Neural Operator (FNO2d) is introduced in the output domain, enabling accurate reconstruction of transient vortex dynamics.

\subsection{FNO-Mamba Module}
The FNO-Mamba module consists of two branches: the \textbf{temporal feature extraction branch}, the \textbf{spatial feature extraction branch}. The former uses the state-space model to extract temporal features from the long-time sensor data, while the latter captures the spatial features of the sensors using a one-dimensional Fourier Neural Operator. The features extracted by both branches are then fused, and high-precision flow field reconstruction output is obtained through iterative computation. 

\subsubsection{Temporal Feature Extraction}
To efficiently capture the historical semantic information contained in the input sequences, the temporal feature extraction branch introduces the Selective State Space Model (SSM)~\cite{mamba} to process the historical input sequences. It primarily includes the following three key steps: (1) \textbf{Selective Input Construction.} Instead of directly feeding raw sensor inputs into the SSM, we first generate control matrices that selectively emphasize important time steps. This is achieved by applying a linear projection followed by a 1D convolution to the input sequence, allowing the model to highlight temporally informative features relevant to physical field dynamics; (2) \textbf{Discretization of Control Matrices in SSM}. Since SSMs originate from continuous-time systems, we discretize the control matrices to adapt them for computation over discrete-time sensor data. A learnable time-step parameter is introduced to improve flexibility and temporal resolution during training; (3) \textbf{Computation of SSM Output.} The SSM updates its hidden state at each time step based on the input and the discretized control matrices. The output at each step is then computed via the observation matrix, resulting in a global temporal representation that summarizes the dynamic evolution across the entire input sequence.

\begin{figure}[h]
	\centering
	\includegraphics[width=0.8\textwidth]{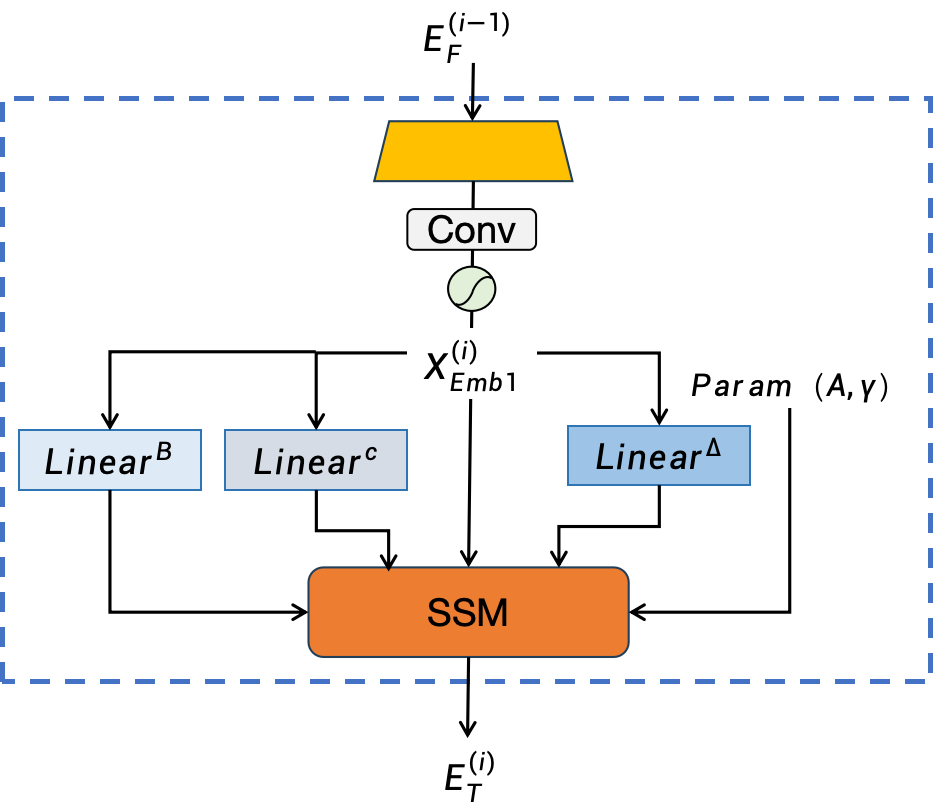}  
	\caption{The overview of temporal feature extraction branch.}
	\label{selective input for ssm}
\end{figure}

{\flushleft\bf Constructing the selective input for SSM.} 
Sensor data often contains redundant temporal information, and the contribution of each time step to the flow field reconstruction may vary significantly. Directly using raw input at every time point can lead to inefficient computation and suboptimal feature utilization. To address this, we construct selective temporal inputs that emphasize informative time steps and suppress irrelevant ones, improving both efficiency and accuracy.

As illustrated in Figure~\ref{selective input for ssm}, this process begins by encoding the input feature from the $(i{-}1)$-th FNO-Mamba block, $E_F^{(i-1)}$, using a linear transformation followed by a 1D convolution. This produces the embedding $X_{\text{Emb1}}^{(i)} \in \mathbb{R}^{B \times L \times M}$, which captures both local and sequential temporal information, where $B$, $L$, and $M$ denote the batch size, sequence length, and feature dimension, respectively.  In parallel, a second embedding $X_{\text{Emb2}}^{(i)}$ is computed from the same input $E_F^{(i-1)}$ for use in the later fusion stage. Specifically, $X_{\text{Emb1}}^{(i)}$ is used to generate SSM-related parameters, while $X_{\text{Emb2}}^{(i)}$ serves as a gating mechanism to adaptively balance the contributions of temporal and spatial features in the final output. The detailed computation of these parameters is as follows:


\begin{align}
    X_{\text{Emb1}}^{(i)} &\in \mathbb{R}^{B \times L \times M} 
    = \mathrm{SiLU}\Big(\mathrm{Conv1d}\big(\mathrm{Linear}(E_F^{(i-1)})\big)\Big), 
    \label{eq:Emb1}\\[6pt]
    X_{\text{Emb2}}^{(i)} &\in \mathbb{R}^{B \times L \times M} 
    = \mathrm{SiLU}\Big(\mathrm{Linear}\big(E_F^{(i-1)}\big)\Big).
\end{align}

Based on $X_{\text{Emb1}}^{(i)}$, three sets of SSM-related parameters are generated using separate linear layers:
\begin{align}
    B^{(i)} &\in \mathbb{R}^{B \times L \times N} 
    = \mathrm{Linear}^B\big(X_{\text{Emb1}}^{(i)}\big), 
    \label{eq:B}\\[6pt]
    C^{(i)} &\in \mathbb{R}^{B \times L \times N} 
    = \mathrm{Linear}^C\big(X_{\text{Emb1}}^{(i)}\big), 
    \label{eq:C}\\[6pt]
    \Delta^{(i)} &\in \mathbb{R}^{B \times L \times M} 
    = log\Bigl(1 + \text{exp}\bigl(\mathrm{Linear}^\Delta\bigl(X_{\text{Emb1}}^{(i)}\bigr) + \gamma\bigr)\Bigr),
    \label{eq:Delta}
\end{align}

where $B^{(i)}, C^{(i)}$ are the input control matrix and observation matrix of the SSM. $\Delta^{(i)}$ is a learnable time-step encoding. $\gamma$ is a learnable bias term. $A$ is the shared state transition matrix initialized as a global trainable parameter. These matrices $(A, B^{(i)}, C^{(i)}, \Delta^{(i)})$ form the full parameter set of the selective SSM, enabling the model to dynamically adapt to each time step and effectively extract task-relevant temporal dynamics. This selective encoding enhances the temporal modeling capacity of the system, especially under sparse or uneven sampling conditions.



{\flushleft\bf Discretization of Control Matrices in SSM.} Since physical field data is typically collected at discrete time intervals, it is necessary to discretize the continuous-time state-space model in order to make it compatible with discrete-time computation. This discretization allows the model to effectively extract temporal features from sequential sensor inputs within a numerically stable and trainable framework.

For the continuous state-space model, the model is discretized using the Zero Order Hold (ZOH) method~\cite{pohlmann2000principles} at a discrete time step \(\Delta\). Let the state transition matrix \(A\) and the input matrix \(B\) of the continuous model be discretized to obtain:
\begin{align}
    \bar{A} \in \mathbb{R}^{B*L*M*N} &= \exp(\Delta\, A), \\
    \bar{B} \in \mathbb{R}^{B*L*M*N} &= A^{-1}\Bigl(\exp(\Delta\, A)-I\Bigr)B,
    \label{eq:combined_formula0}
\end{align}
where \(I\) is the identity matrix. The discretized state transition matrix \(\bar{A}\) and the input matrix \(\bar{B}\) are suitable for the discrete representation of transient physical fields and are used to extract and represent historical temporal features at each time step. 

{\flushleft\bf Computation of SSM Output.} After completing the discretization of the SSM input (i.e., obtaining parameters such as \(\bar{A}\), \(\bar{B}\), and \(C^{(i)}\)), the discrete state-space model is used to iteratively update the sequence's hidden states. The update equations are defined as follows:

\begin{align}
\label{eq:SSM01}
h(t_k) &= \bar{A}\, h(t_{k-1}) + \bar{B}\, x(t_k), \\
y(t_k) &= C^{(i)}\, h(t_k),
\end{align}

where \(t_k \in \{1,2,\dots,L\}\) denotes the time step, \(h(t_k) \in \mathbb{R}^{N}\) is the hidden state at time \(t_k\), and \(x(t_k)\) is the input signal at the corresponding time step. 

\begin{figure}[t]
	\centering
	\includegraphics[width=0.95\textwidth]{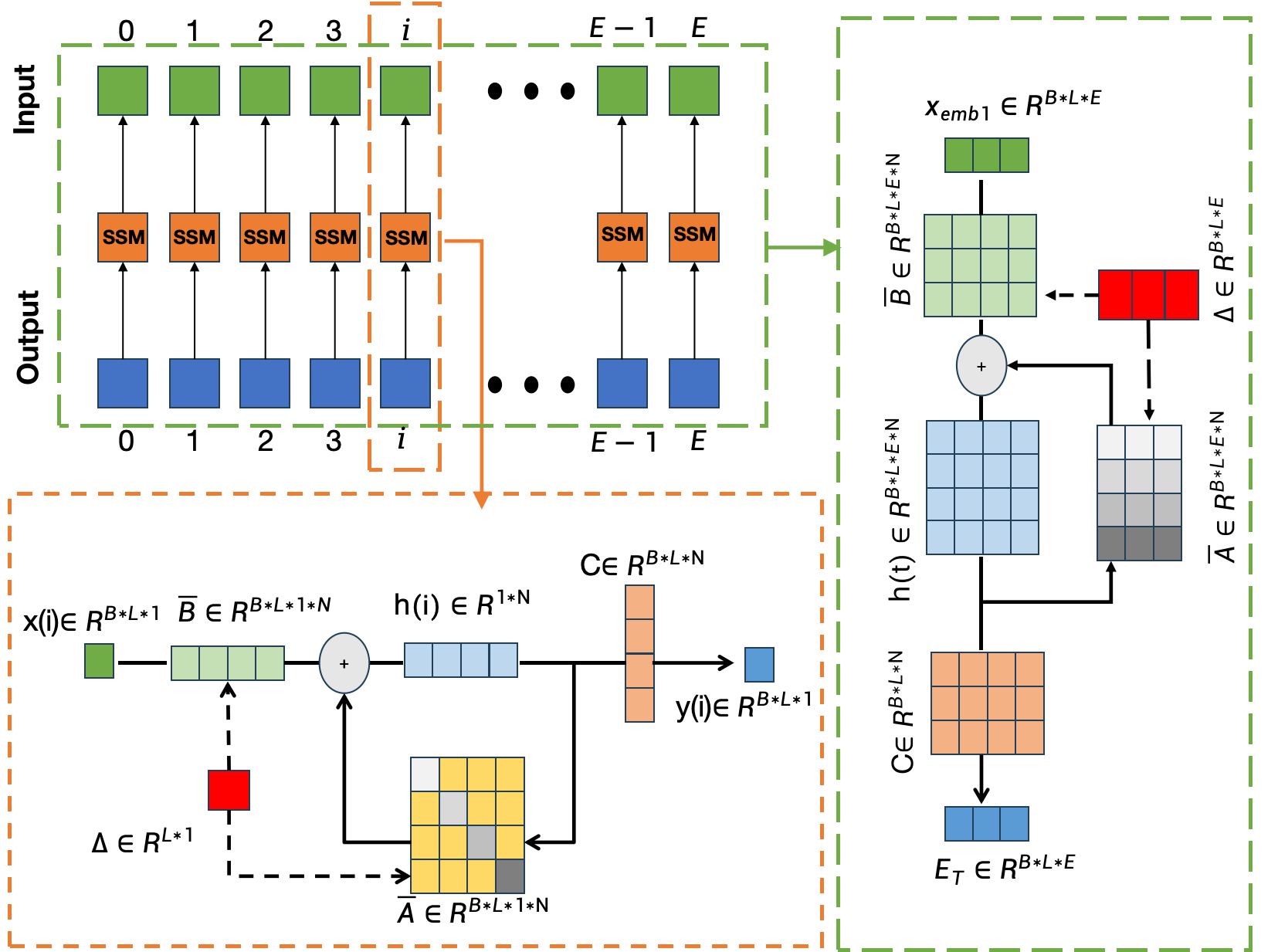}  
	\caption{The overall calculation process of SSM.}
	\label{fig:The overall calculation process of SSM}
\end{figure}

As shown in Figure~\ref{fig:The overall calculation process of SSM}, the entire SSM computation process can be viewed as a parallel state compression performed independently across each feature dimension. After updating all time steps, the final hidden state is mapped to the target feature space using the matrix \(C^{(i)}\), resulting in the temporal feature \(E_T^{(i)}\) that integrates historical information. Additionally, this iterative process can also be viewed as a global convolution operation:

\begin{equation}
\begin{aligned}
    K &= \bigl(C\overline{B},\, C\overline{A}\,\overline{B},\, \dots,\, C\overline{A}^L\,\overline{B},\, \dots\bigr),\\[1ex]
    y &= x \cdot \overline{K}.
\end{aligned}
\end{equation}

Where \(\overline{K} \in \mathbb{R}^L\) can be considered as a global convolution kernel. By utilizing the SSM, the temporal feature extraction branch efficiently captures long-term temporal dependencies in historical sensor observations with linear time complexity. However, the independent modeling strategy for each feature dimension in the SSM neglects the inherent spatial relationships within the sensor observation inputs. Therefore, to effectively integrate spatial features, another spatial feature extraction network is required to achieve this task.

\subsubsection{Spatial Feature Extraction}
To address the limitations of the temporal feature extraction branch in capturing spatial information from sensors, we construct a spatial feature extraction branch to capture spatial information. It uses the Fourier Neural Operator (FNO1d) to map the original features to the frequency domain and models the global spatial receptive field through frequency-domain convolution, effectively capturing the spatial position information embedded in the sensor observations and enhancing the extraction of global spatial features. 

\begin{figure}[t]
	\centering	\includegraphics[width=0.95\textwidth]{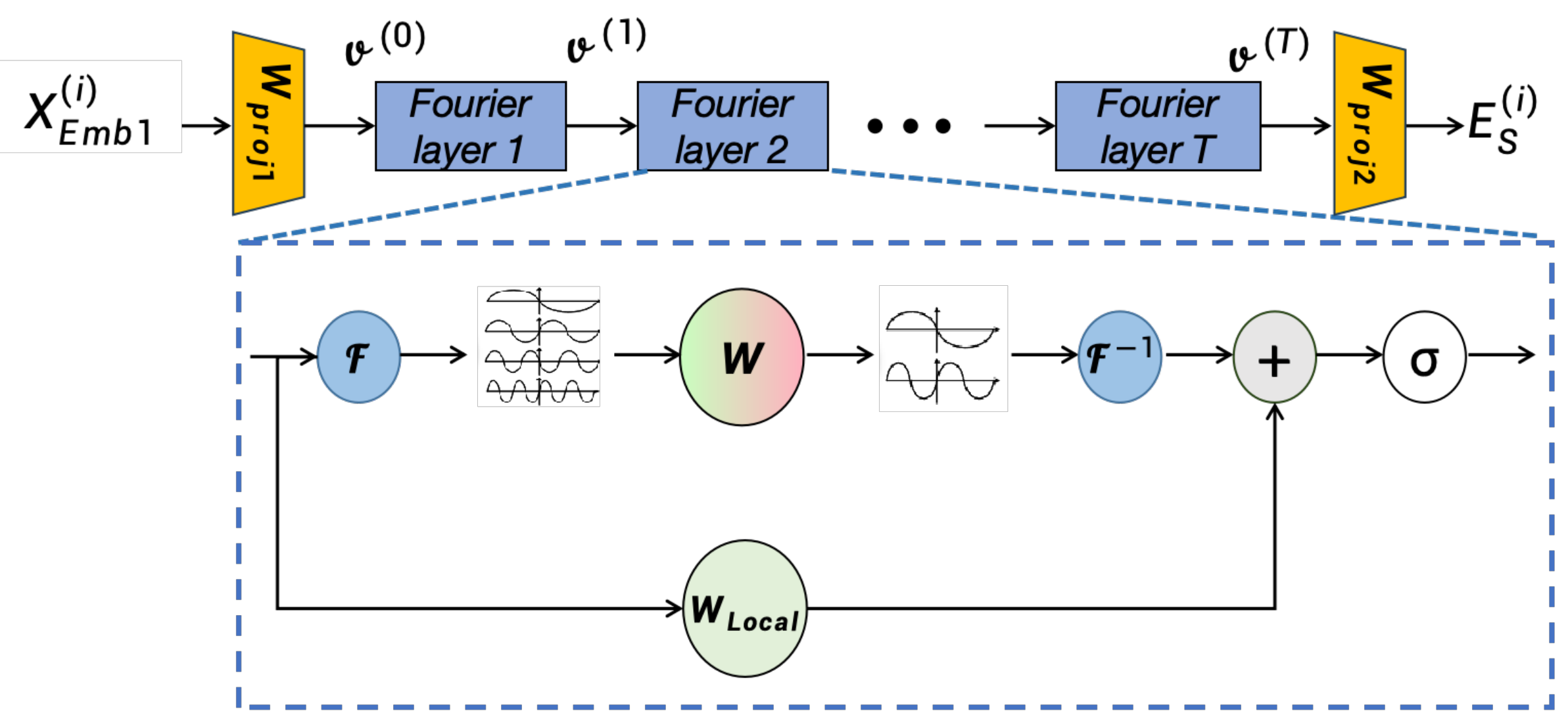}  
	\caption{The overview of spatial feature extraction branch.}
    \label{fig:overview of spatial feature extraction branch}
\end{figure}

As shown in the Figure~\ref{fig:overview of spatial feature extraction branch}, the initial embedding of the spatial feature branch is denoted as $X_{\text{Emb1}}^{(i)} \in \mathbb{R}^{B \times L \times M}$, where $B$ is the batch size, $L$ represents the sequence length (i.e., the number of time steps), and $M$ denotes the feature dimension. For clarity, we omit the batch dimension in the following derivations and focus on the computation for a single sample, treating $X_{\text{Emb1}}^{(i)} \in \mathbb{R}^{L \times M}$. For each time step $j \in [1, L]$, the corresponding input 
$X_{\text{Emb1},j}^{(i)} \in \mathbb{R}^{1 \times M}$ can be regarded as a single-channel 
$M$-dimensional spatial feature function. We first project this signal into a higher-dimensional feature space using a linear mapping:

\begin{equation}
v_j^{(0)} \in \mathbb{R}^{M \times d} = W_{\text{proj1}}\, X_{\text{Emb1},j}^{(i)},
\end{equation}

where \(W_{\text{proj1}}\) is the projection matrix. $d$ denotes the internal feature dimension in the Fourier layers, used to expand the spatial token representation for frequency-domain modeling. Next, for the transformed features \(\{v_j^{(0)}\}_{j=1}^{L}\), we use \(T\) layers (\(T=4\)) of Fourier layers for iterative computation to further extract spatial features from the input data. Let the input of the \(\ell\)-th layer be \(v_j^{(\ell)} \in \mathbb{R}^{M \times d}\) 
(where \(\ell=0,1,\dots,T-1\)). The computational process consists of the following steps:

For each channel, the input signal is first transformed into the frequency domain using a one-dimensional Fourier transform \(\mathcal{F}\):

\begin{equation}
\hat{v}_j^\ell \;=\; \mathcal{F}\bigl(v_j^\ell\bigr),
\end{equation}

where the Fourier transform \(\mathcal{F}\) is implemented using the one-dimensional Discrete Fourier Transform (DFT). The specific calculation is given by:

\begin{equation}
\hat{v}_j^\ell[k,c] \;=\; \sum_{x=0}^{M-1} v_j^\ell[x,c]\,
e^{-2\pi i\,\frac{kx}{M}}, 
\quad k = 0,1,\ldots,M-1.
\end{equation}

To accelerate computation, only the first \(m\) frequency components are retained, where \(m \ll M\), as high-frequency components are typically associated with transient variations in physical fields, which can be captured by the temporal branch.

For each retained frequency \(k \in \{0,1,\dots,m-1\}\), a learnable complex-valued weight matrix 
\(W_{k} \in \mathbb{C}^{d \times d}\) is applied to perform weighting, learning the globally intrinsic positional information and enhancing global spatial feature fusion. The updated frequency-domain features \(\hat{u}_j^{\ell}\) are defined as:

\begin{equation}
    \hat{u}_j^{\ell}[k,c'] = \sum_{c=1}^{d} \hat{v}_j^{\ell}[k,c]\, W_{k}(c,c'), \quad c' = 1, \dots, d.
\end{equation}

The weighted frequency-domain representation is then transformed back to the spatial domain using the inverse one-dimensional Fourier transform \(\mathcal{F}^{-1}\):

\begin{equation}
\tilde{v}_j^\ell \;=\; \mathcal{F}^{-1}\bigl(\hat{u}_j^\ell\bigr).
\end{equation}

The inverse Fourier transform \(\mathcal{F}^{-1}\) is implemented using the one-dimensional inverse Discrete Fourier Transform (IDFT), which is computed as:

\begin{equation}
\tilde{v}_j^\ell[x,c']
\;=\; \frac{1}{d}\sum_{k=0}^{m-1} \tilde{u}_j^\ell[k,c']\,
e^{\,2\pi i\,\frac{kx}{M}},
\quad x = 0,1,\ldots,M-1.
\end{equation}

A local linear transformation \(W_{\text{local}}\) is applied to the inverse-transformed result \(\tilde{v}_j^\ell\) and the original local feature \(v_j^\ell\), followed by a nonlinear activation function \(\sigma(\cdot)\) to update the output:

\begin{equation}
v_j^{\ell+1}(x) = \sigma\Bigl(W_{\text{local}}\, v_j^{\ell}(x) + \tilde{v}_j^{\ell}(x)\Bigr),
\end{equation}

where \(x\) denotes the spatial feature position index. The above core operations can be viewed as a discretized form of the kernel integral operator \(\mathcal{K}\) in the spatial domain, which can be expressed in the continuous form as:

\begin{equation}
 \label{eq:fno}
    \bigl(\mathcal{K} v_j^{\ell}\bigr)(x) = \int k(x,y)\, v_j^{\ell}(y)\,dy.
\end{equation}

The discrete implementation utilizes the Fourier transform to convert the integral operation of the kernel integral operator \(\mathcal{K}\) in Eq. \eqref{eq:fno} into weighted summation operations in the frequency domain. During this process, updating the parameter \(W\) in the frequency domain effectively realizes the function of the convolution kernel \(\mathcal{K}\). Subsequently, the frequency-domain results are mapped back to the spatial domain through the inverse Fourier transform, thereby constructing a global receptive field covering the entire input space and achieving the extraction of intrinsic spatial features.

After $T$ iterative layers, the final output is mapped back to the original input dimension using a linear layer:

\begin{equation}
E_S^{(i)} = W_{\text{proj2}}\left(\{v_j^{(T)}\}_{j=1}^{L}\right).
\end{equation}

\subsection{Flow Reconstruction Pipeline}
\subsubsection{Feature Fusion Between Temporal and Spatial Branches.}
\begin{figure}[h]
	\centering
	\includegraphics[width=0.95\textwidth]{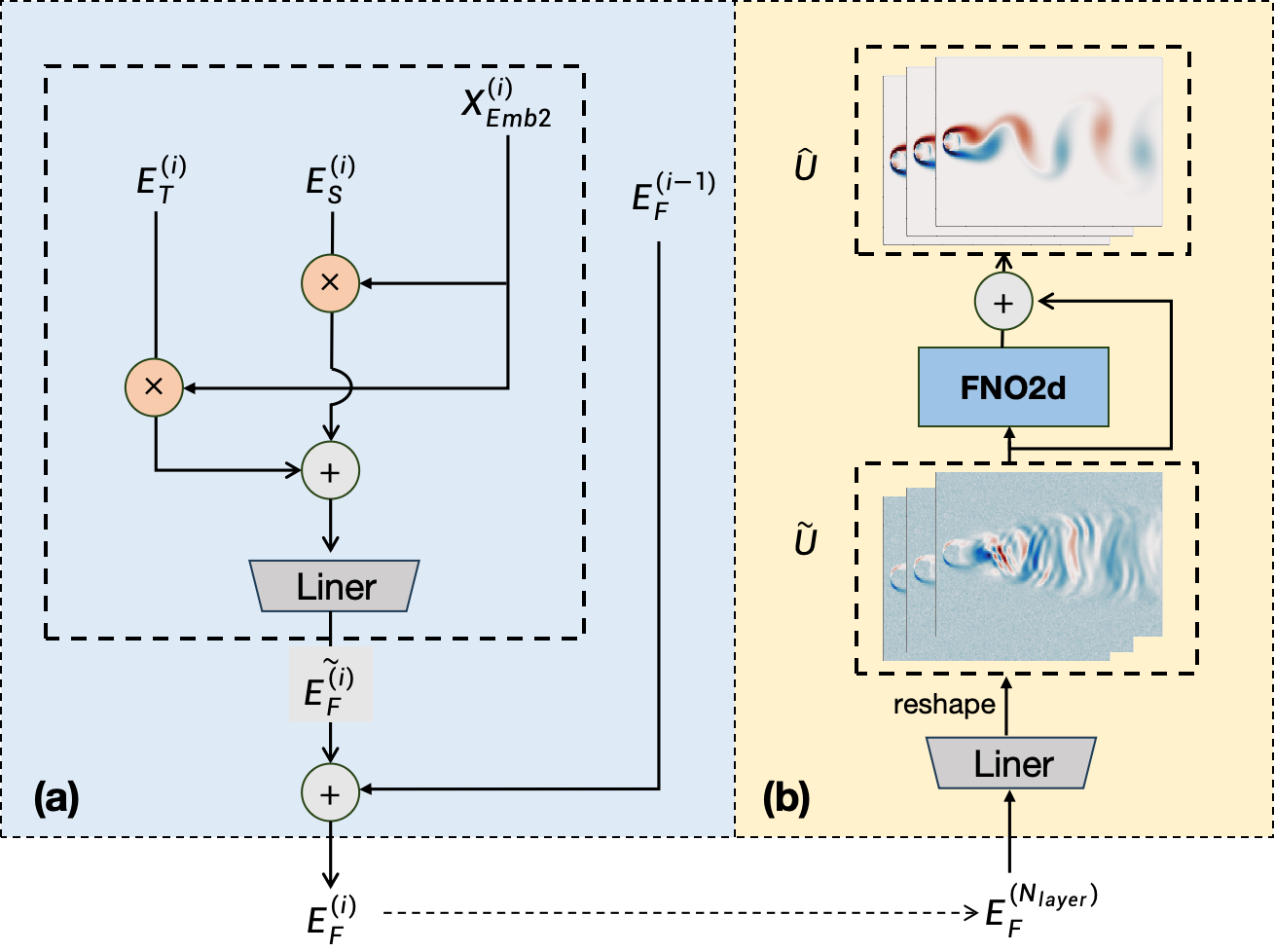}  
	\caption{The feature fusion of our FR-Mamba.}
	\label{fig:feature fusion}
\end{figure}
As shown in Figure~\ref{fig:feature fusion}(a), the temporal and spatial feature extraction branches in the $i$-th FNO-Mamba block generate temporal and spatial features, $E_T^{(i)}$ and $E_S^{(i)}$, respectively. To effectively fuse these features, a gating mechanism is introduced, using an auxiliary embedding $X_{\text{Emb2}}^{(i)}$ to dynamically weight their contributions:

\begin{align}
    \tilde{E}_F^{(i)} &= X_{\text{Emb2}}^{(i)} \odot E_T^{(i)} + X_{\text{Emb2}}^{(i)} \odot E_S^{(i)}, \\
    E_F^{(i)} &=  \tilde{E}_F^{(i)} + E_F^{(i-1)},
\end{align}

where \(\odot\) denotes the Hadamard product. This design enables adaptive adjustment of spatiotemporal feature importance for better transient physics representation.

\subsubsection{Flow Field Reconstruction}
As shown in Figure \ref{fig:feature fusion}(b), after \(N_{layer}\) iterations, the final fused spatiotemporal feature \(E_F^{(N_{layer})} \in \mathbb{R}^{B \times L \times d}\) is obtained. A feedforward network (FFN) maps this feature to the target flow field dimension:

\begin{equation}
    \tilde{U} = \operatorname{reshape}\bigl(\operatorname{FFN}(E_F^{(N_{layer})})\bigr).
\end{equation}

Here, the output of the FFN is reshaped from $\mathbb{R}^{B \times L \times C}$ to $\mathbb{R}^{B \times H \times W \times C}$, where $L = H \times W$ corresponds to the spatial resolution of the reconstructed flow field, and $C$ is the number of output channels (e.g., velocity components).
To enhance global spatial feature fusion, a 2D Fourier operator with residual connection refines the reconstruction:

\begin{equation}
    \hat{U} = \tilde{U} + \operatorname{FNO2d}(\tilde{U}),
\end{equation}
where the 2D Fourier operator is an extension of the 1D Fourier operator to two-dimensional spatial inputs.

\section{Experiments}

\subsection{Experimental Setup}
\subsubsection{Datasets and Metrices}
\begin{figure}[h]
	\centering
\includegraphics[width=0.7\textwidth]{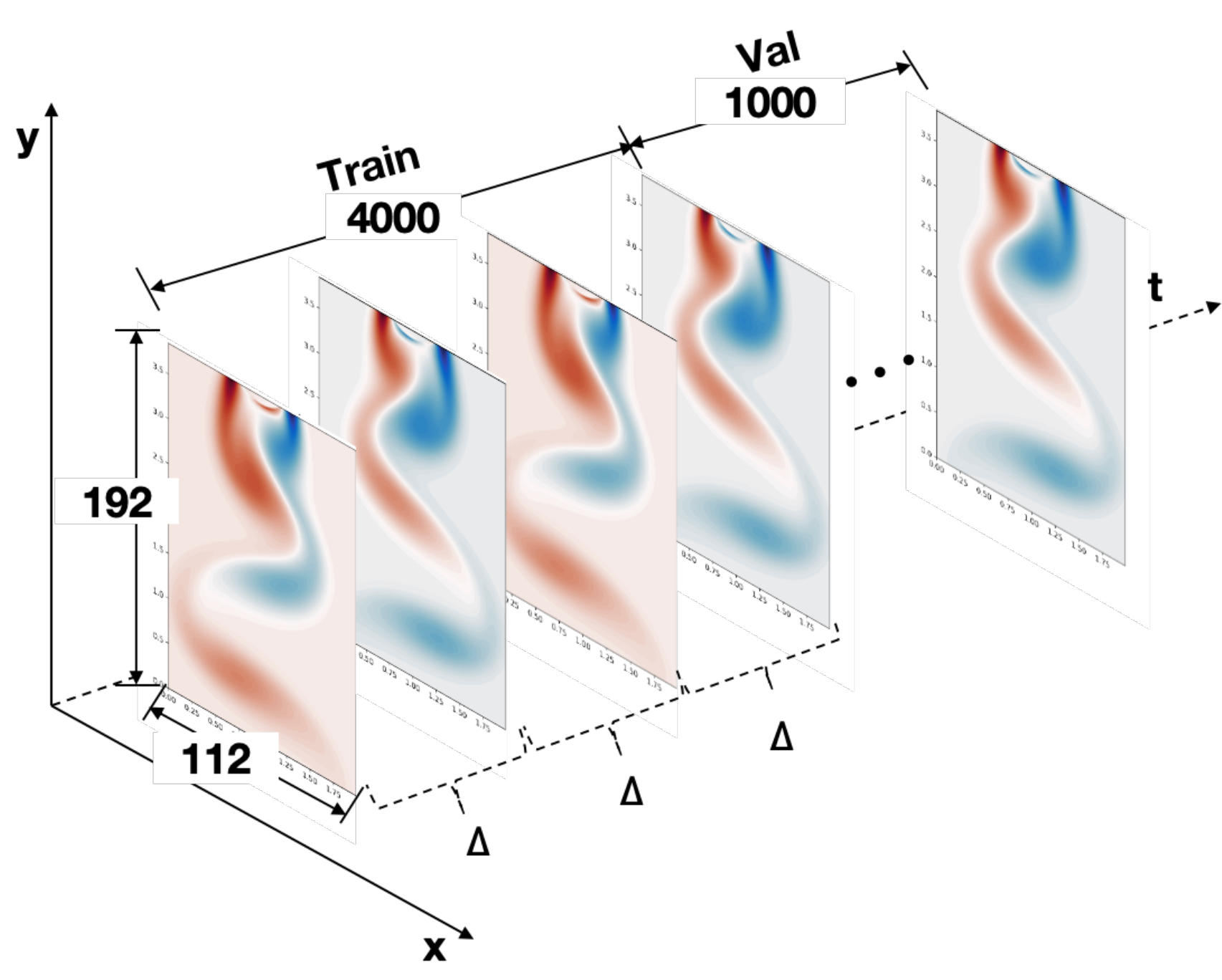}  
	\caption{Structure of the time-resolved flow field dataset with 4000 training and 1000 validation snapshots.}
	\label{fig:data}
\end{figure}

We evaluate our method on the same two-dimensional unsteady flow dataset as used in~\cite{VoronoiCNN, GappyMLP}, which simulates flow around a circular cylinder at a Reynolds number of $Re = 100$.
The dataset consists of observation data collected by 16 sensors uniformly distributed in the two-dimensional flow field. These observation data are used to reconstruct the global vortex field, thereby verifying the model's performance in reconstructing time-series physical fields over long time sequences. Time-series snapshots of the vortex field are obtained by solving the incompressible Navier-Stokes equations, as described by the following equation:

\begin{equation}
\frac{\partial u_i}{\partial t} + u_j \frac{\partial u_i}{\partial x_j} = -\frac{1}{\rho} \frac{\partial p}{\partial x_i} + \nu \frac{\partial^2 u_i}{\partial x_j \partial x_j},
\end{equation}

where $u_i$ and $p$ represent velocity and pressure, respectively, and $\nu$ is the dynamic viscosity of the fluid. The resolution of each vortex field snapshot is (192,112), represented by uniformly discretized two-dimensional images. The sampling time interval between snapshots in the dataset is $\Delta$, and it contains a total of 5000 temporally consecutive transient vortex field snapshots, covering multiple vortex shedding cycles. In the experiments, the first 400 snapshots are used for training, and the last 1000 snapshots are used for testing, as illustrated in Figure~\ref{fig:data}. Within the grid region of the flow field, 16 sensors are uniformly distributed to collect measurements as the model input. The ultimate goal of the model is to reconstruct the transient vortex field based on these time-sequenced discrete sensor input data. For metrics, we use Mean Absolute Error (MAE) and Maximum Absolute Error (Max-AE) as our evaluation metrics.

\subsubsection{Loss Functions}

We use the mean absolute error between the reconstructed vortex field sequence $\hat{U}$ and the true vortex field sequence $U$ as the loss function. It is defined as follows:
\begin{equation}
    \mathcal{L} = \frac{1}{ML} \sum_{l=1}^{L}\sum_{m=1}^{M} \left| \hat{U}_{l,m} - U_{l,m} \right|,
\end{equation}
where $M$ is the total number of discrete points in the spatial domain of the target physical field, and $L$ is the number of discrete vortex field snapshots in the transient vortex field sequence. This process is treated as a regression task, where the model parameters are updated by minimizing the mean absolute error between the true and reconstructed physical fields.

\subsection{Main Results}
\subsubsection{Quantitative comparison with other methods}
As shown in Table~\ref{table:cylinder_mae_results} and Table~\ref{table:cylinder_max_ae_results}, we quantitatively compare various physical reconstruction methods on time-evolving flow field dataset. As shown,  FR-Mamba
achieves the lowest reconstruction error on the task of flow around a circular cylinder,
manifesting its superiority over existing methods including SVR-POD~\cite{SVR-POD}, LSTM-POD~\cite{LSTM}, Voronoi UNet~\cite{Voronoi-UNet}, Shallow Decoder~\cite{shallowdecoder}, Voronoi CNN~\cite{VoronoiCNN}, Gappy MLP~\cite{GappyMLP}, and PFR-Transformer~\cite{PFR-Transformer}, on the metrices of mean absolute error (MAE) and maximum absolute error (Max-AE). To further assess performance during different vortex evolution stages, four time intervals (T1: 3001–3400, T2: 3401–3800, T3: 3801–4200, T4: 4201–4600, T5: 4601-5000) are also examined.

\begin{table}[t]
\centering
\resizebox{\textwidth}{!}{
\begin{tabular}{@{}cccccccc@{}}
\toprule
\multirow{2}{*}{Method} & \multirow{2}{*}{Backbones} & \multicolumn{6}{c}{Mean Absolute Error (e-4)} \\ \cmidrule(l){3-8}
                         &           & T1 & T2 & T3 & T4 & T5 & Ave. \\ \midrule
SVR-POD~\cite{SVR-POD}           & SVR         & 27.98 & 33.36 & 38.05 & 41.96 & 55.35 & 39.33 \\
LSTM-POD~\cite{LSTM}             & LSTM        & 1.36  & 1.42  & 1.44  & 1.48  & 1.50  & 1.44 \\
Shallow Decoder~\cite{shallowdecoder} & FCN         & 3.18  & 3.54  & 3.66  & 3.81  & 4.11  & 3.66 \\
Voronoi CNN~\cite{VoronoiCNN}         & CNN         & 14.74 & 23.04 & 24.74 & 25.00 & 38.16 & 25.14 \\
Voronoi UNet~\cite{Voronoi-UNet}      & UNet        & 1.62  & 1.68  & 1.70  & 1.71  & 1.72  & 1.69 \\
Gappy MLP~\cite{GappyMLP}             & MLP         & 3.16  & 3.52  & 3.54  & 3.65  & 3.81  & 3.54 \\
PFR-Transformer~\cite{PFR-Transformer} & Transformer & 0.84 & 0.86 & 0.95 & 0.96 & 0.97 & 0.91 \\
\textbf{FR-Mamba}                     & \textbf{Mamba} & \textbf{0.30} & \textbf{0.31} & \textbf{0.31} & \textbf{0.32} & \textbf{0.32} & \textbf{0.31} \\ \bottomrule
\end{tabular}
}
\caption{Comparison of Mean Absolute Error (MAE) for various methods in reconstructing time-series cylinder flow. T1–T5 represent different time intervals.}
\label{table:cylinder_mae_results}
\end{table}

Results show that FR-Mamba consistently achieves the lowest reconstruction error across all intervals, with MAE ranging from 0.30 to 0.32 and Max-AE remaining around 0.10, demonstrating both accuracy and stability.  In contrast, Voronoi UNet~\cite{Voronoi-UNet} achieves a higher MAE of 1.69 and Max-AE of 0.74, despite benefiting from convolutional structures and skip connections to better capture multi-scale features. However, it requires additional sensor location metadata and image-format conversion, which increases implementation complexity. Shallow Decoder~\cite{shallowdecoder} and Gappy MLP~\cite{GappyMLP}, both based on fully connected networks, achieve MAEs of 3.66 and 3.54 respectively. While they improve upon traditional baselines like SVR-POD~\cite{SVR-POD} (MAE: 39.33), they do not explicitly incorporate sensor position information, which limits their spatial generalization. Gappy MLP further incorporates historical flow priors to enhance temporal constraints, but still suffers from Max-AE as high as 1.19, and is highly sensitive to sensor placement—struggling to reconstruct complex, high-dimensional spatiotemporal patterns.

Overall, FR-Mamba combines the advantages of prior methods while eliminating their key limitations, achieving accurate and robust reconstruction of the time-series flow field reconstruction. 

\begin{table}[t]
\centering
\resizebox{\textwidth}{!}{
\begin{tabular}{@{}cccccccc@{}}
\toprule
\multirow{2}{*}{Method} & \multirow{2}{*}{Backbones} & \multicolumn{6}{c}{Max Absolute Error (e-2)} \\ \cmidrule(l){3-8}
                         &           & T1 & T2 & T3 & T4 & T5 & Ave. \\ \midrule
SVR-POD~\cite{SVR-POD}           & SVR         & 25.38 & 38.22 & 52.00 & 55.37 & 65.33 & 47.26 \\
LSTM-POD~\cite{LSTM}             & LSTM        & 0.72  & 0.76  & 0.90  & 0.92  & 1.09  & 0.88 \\
Shallow Decoder~\cite{shallowdecoder} & FCN         & 1.11  & 1.26  & 1.37  & 1.41  & 1.68  & 1.37 \\
Voronoi CNN~\cite{VoronoiCNN}         & CNN         & 10.45 & 12.94 & 13.50 & 14.13 & 15.23 & 13.25 \\
Voronoi UNet~\cite{Voronoi-UNet}      & UNet        & 0.67  & 0.71  & 0.72  & 0.74  & 0.84  & 0.74 \\
Gappy MLP~\cite{GappyMLP}             & MLP         & 0.98  & 1.12  & 1.19  & 1.21  & 1.45  & 1.19 \\
PFR-Transformer~\cite{PFR-Transformer} & Transformer & 0.49 & 0.52 & 0.54 & 0.54 & 0.58 & 0.53 \\
\textbf{FR-Mamba}                     & \textbf{Mamba} & \textbf{0.10} & \textbf{0.10} & \textbf{0.10} & \textbf{0.10} & \textbf{0.11} & \textbf{0.10} \\ \bottomrule
\end{tabular}
}
\caption{Comparison of Max Absolute Error (Max-AE) for various methods in reconstructing time-series cylinder flow. T1–T5 represent different time intervals.}
\label{table:cylinder_max_ae_results}
\end{table}

\subsubsection{Quantitative comparison in different regions}
\begin{figure}[t]
	\centering
	\includegraphics[width=0.95\textwidth]{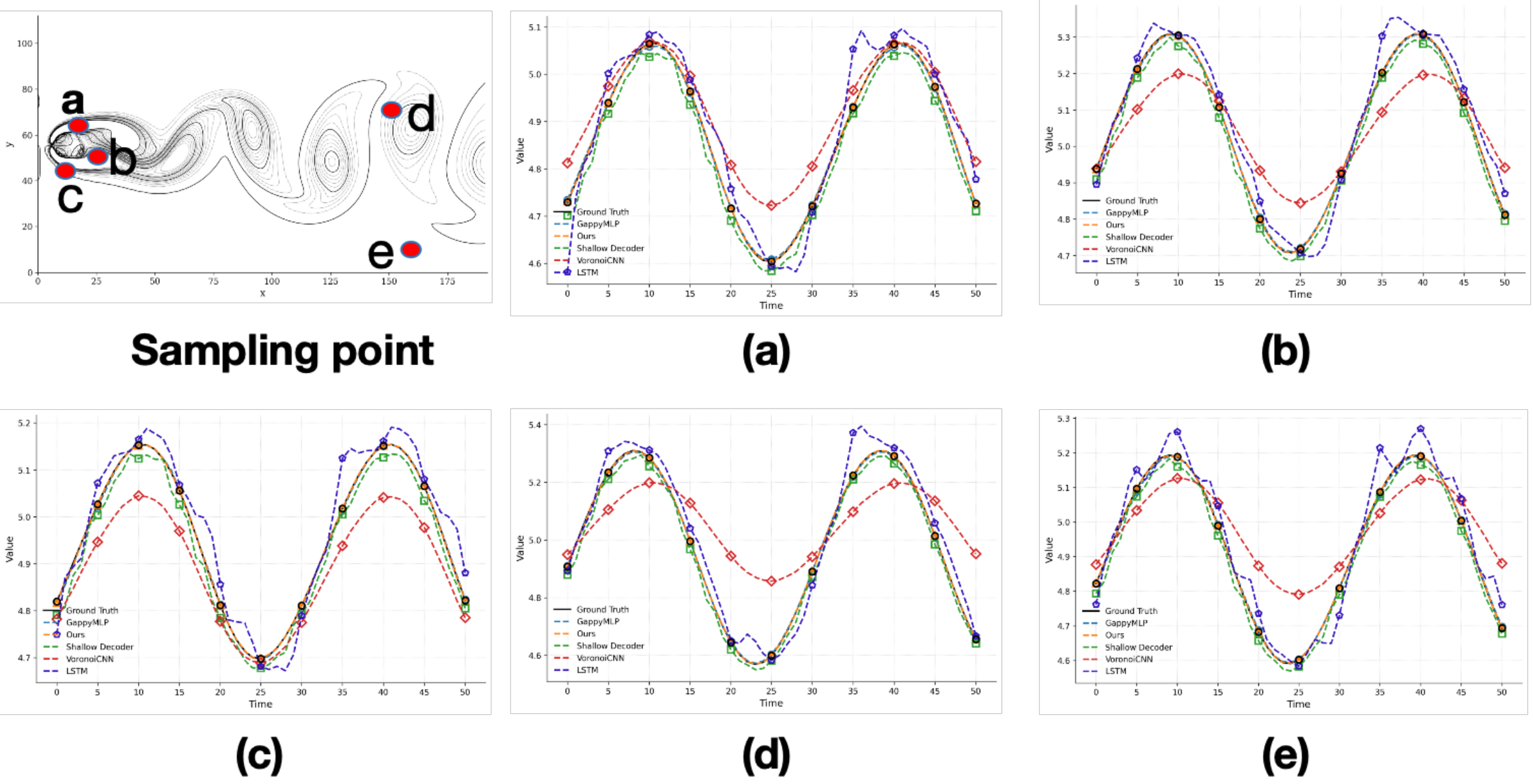}  
	\caption{Visualization of different regions in the flow field.}
	\label{fig:different regions}
\end{figure}

To further evaluate the proposed model's temporal reconstruction performance across these regions, Figure~\ref{fig:different regions} presents visualizations of time-series flow field predictions generated by different methods. During the flow around a cylinder, the flow field can be divided into several distinct regions, including the near-wall region, wake region, and far-field region, each exhibiting different flow characteristics and evolution patterns. 
Specifically, Figure~\ref{fig:different regions}(a) illustrates five sampling points: three randomly selected from the near-wall region (points a, b, and c), one from the wake region (point d), and one from the far field (point e). The vorticity values at these locations were tracked over 50 consecutive time steps. Figures~\ref{fig:different regions}(b) to \ref{fig:different regions}(f) show the reconstruction results from different methods—Gappy MLP, Shallow Decoder, Voronoi CNN, and the proposed FR-Mamba—with the black curve representing the ground truth.

As shown in these visualizations, differences in vorticity dynamics across regions lead to varying reconstruction performance. For instance, Shallow Decoder achieves relatively low error in the wake and far-field regions between time steps 30 to 40 but exhibits noticeably higher errors in the near-wall region during the same period. This highlights the difficulty existing methods face when modeling complex, high-dimensional spatiotemporal dynamics—particularly in regions with sharp vorticity gradients—resulting in degraded reconstruction quality.

In contrast, the proposed FR-Mamba performs consistently well across all regions and time steps. It accurately fits the true time-varying vorticity curves for all sampling points, including those in regions with sharp changes. Notably, FR-Mamba effectively captures the complex evolution patterns even in highly dynamic regions, enabling high-precision temporal reconstruction.

\begin{figure}[htbp]
	\centering	\includegraphics[width=0.95     \textwidth]{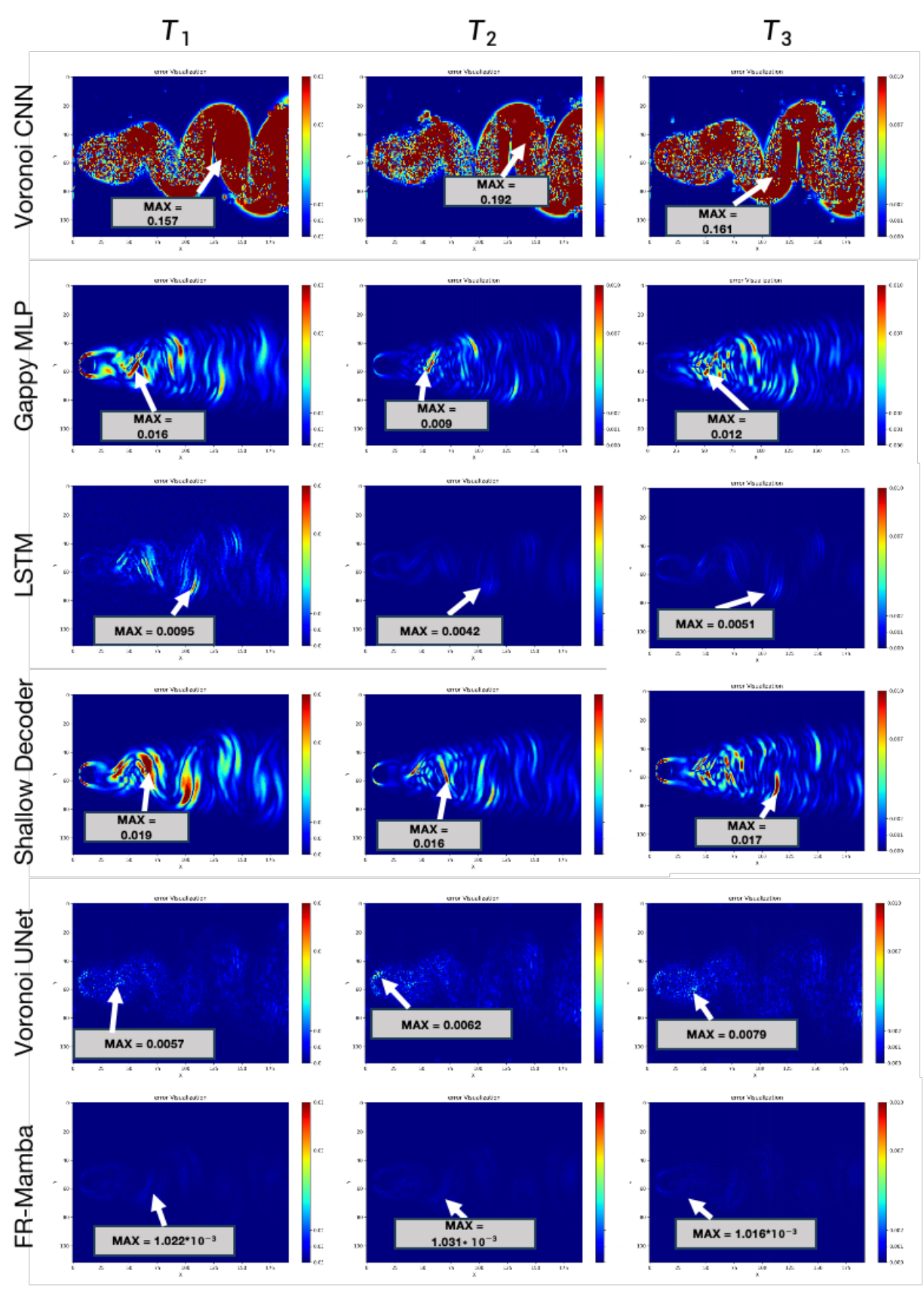}  
	\caption{The qualitative comparison of our FR-Mamba with other representative methods.}
	\label{fig:qualitative Comparison with other works}
\end{figure}

\begin{figure}[htbp]
	\centering	\includegraphics[width=0.95     \textwidth]{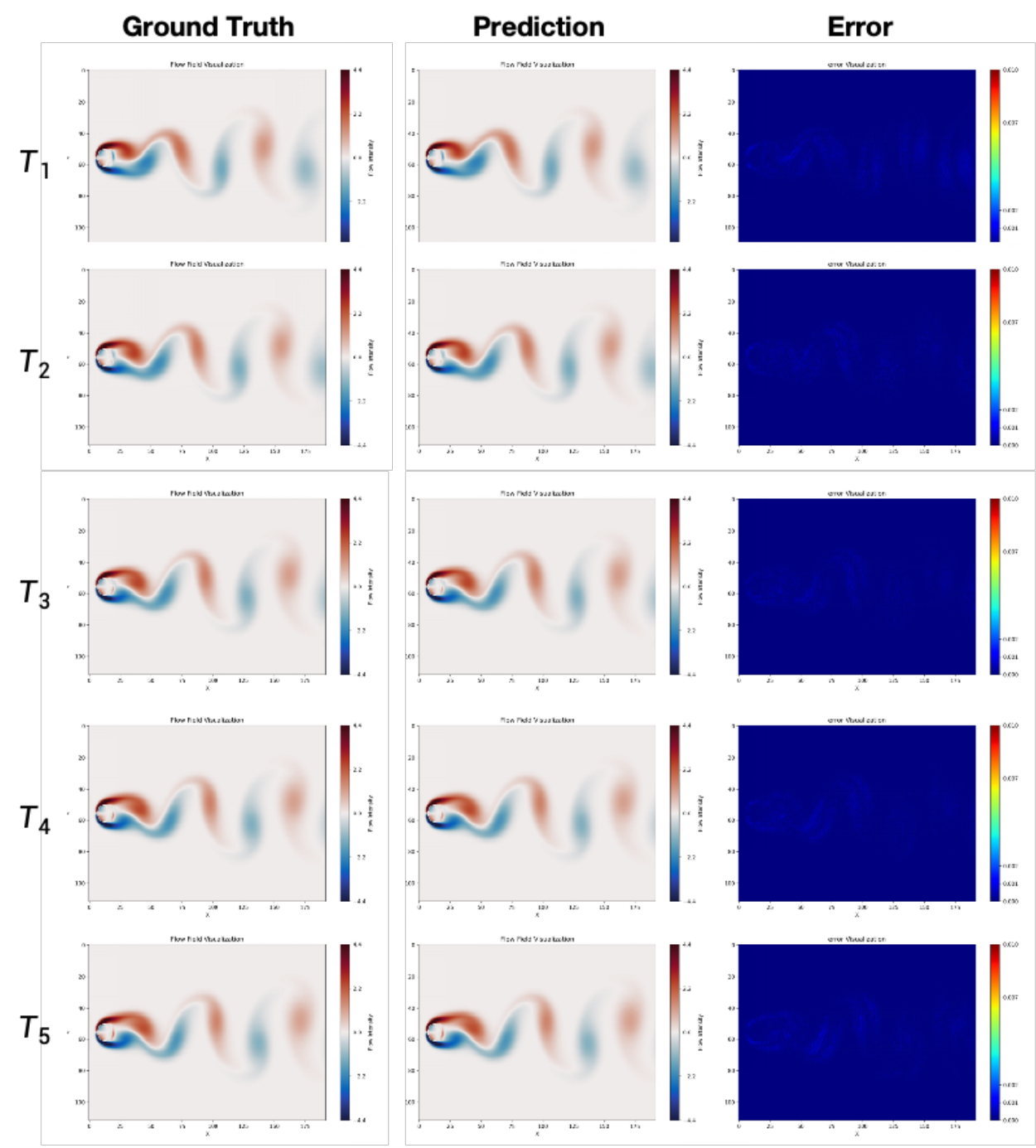}  
	\caption{Visualization of prediction results and corresponding error of our FR-Mamba.}
	\label{fig:Visualization of prediction results}
\end{figure}

\subsubsection{Qualitative Comparison with Other Methods}
Figure~\ref{fig:qualitative Comparison with other works} presents a qualitative comparison of reconstruction error maps at three representative time steps ($T1$–$T3$), selected from the test set. We compare six methods: Gappy MLP~\cite{GappyMLP}, LSTM~\cite{LSTM}, Shallow Decoder~\cite{shallowdecoder}, Voronoi CNN~\cite{VoronoiCNN}, Voronoi UNet~\cite{Unet}, and the proposed FR-Mamba. From the results, FR-Mamba achieves the lowest reconstruction error across all time steps, significantly reducing the global maximum error from 1.170 (Gappy MLP) to \(1.02 \times 10^{-3}\). While Shallow Decoder performs reasonably well in smooth regions, it exhibits large errors in areas with sharp flow transitions, indicating limited capacity for modeling complex spatial structures. Voronoi UNet improves reconstruction in these regions by leveraging its multi-scale convolutional architecture, but it requires additional sensor layout information and image-format inputs. Gappy MLP introduces temporal priors to enhance reconstruction and reduces the peak error to 0.016 and 0.009 at \(T_1\) and \(T_2\), respectively. However, its global error remains high under complex dynamics. These results highlight FR-Mamba’ effectiveness in reconstructing complex, time-evolving physical fields.

\subsubsection{Visualization of Error on Different Time Sequences}
As shown in Figure~\ref{fig:Visualization of prediction results}, to evaluate the performance of the proposed FR-Mamba model on the vortex field reconstruction task, five time steps \((T_1 - T_5)\) were randomly selected from the test set for visual comparison. Each row displays the ground truth vortex snapshot, the reconstructed result by the model, and the corresponding absolute error distribution. It can be observed that FR-Mamba accurately reconstructs the major spatial structures of the vortex field and consistently maintains low reconstruction errors across different time steps, demonstrating strong reconstruction accuracy and robust spatiotemporal generalization capability.

\begin{table}[t]
\centering
\resizebox{\textwidth}{!}{
\begin{tabular}{ccccccc}
\toprule
\multirow{2}{*}{\(N_{layer}\)} & \multicolumn{2}{c}{D = 8}           & \multicolumn{2}{c}{D = 16}           & \multicolumn{2}{c}{D = 32}           \\ \cmidrule(l){2-7}
                    & MAE(e-4)          & Max-AE(e-2)      & MAE(e-4)          & Max-AE(e-2)      & MAE(e-4)          & Max-AE(e-2)      \\ \midrule
5                   & 0.411              & 0.229             & 0.382              & 0.221             & 0.375             & 0.202             \\
10                  & \textbf{0.310}              & \textbf{0.108}             & 0.332              & 0.173            & 0.405              & 0.221            \\
20                  & 0.510             & 0.273             & 0.553             & 0.299             & 0.625             & 0.305            \\ \bottomrule
\end{tabular}
}
\caption{Varing of FNO-Mamba layers and hidden state dimension.}
\label{tb:超参数消融结果}
\end{table}

\begin{table}[t]
	\centering
	\begin{tabular}{cccc}
		\toprule
		FNO1d & FNO2d & MAE(e-4) & Max-AE(e-2) \\ \midrule
		\textbf{--} & \textbf{--} & 0.397 & 0.355 \\
		\checkmark & -- & 0.315 & 0.125 \\
		\checkmark& \checkmark & \textbf{0.310} & \textbf{0.108} \\
		\bottomrule
	\end{tabular}
	\caption{The ablation studies on  1D and 2D Fourier Neural Operators.}
	\label{table:ablation2}
\end{table}

\subsection{Ablation Study on Network Depth and Hidden Dimension}
As shown in Table~\ref{tb:超参数消融结果}, we conduct an ablation study by varying the number of FNO-Mamba layers ($N_{\text{layer}}$) and the hidden state dimension ($D$). When the model is shallow ($N_{\text{layer}} = 5$), increasing $D$ from 8 to 32 leads to consistent improvements in both MAE and Max-AE, indicating enhanced capacity to capture transient physical dynamics. At a moderate depth ($N_{\text{layer}} = 10$), the model achieves its best overall performance with $D = 8$, reaching a MAE of 0.310 and Max-AE of 0.108. Increasing $D$ beyond this point results in diminished gains or slight degradation, suggesting a trade-off between model capacity and optimization stability. However, when the number of layers increases to 20, performance drops significantly across all $D$ values. This indicates that excessive depth leads to parameter redundancy and overfitting, especially under limited training data. Overall, a moderate depth with a compact hidden dimension yields the best trade-off between accuracy and efficiency for physical field reconstruction.
\subsection{Ablation Study on the Fourier Neural Operator}

As shown in Table~\ref{table:ablation2}, we perform ablation studies on the 1D Fourier Neural Operator (FNO1d) in the FNO-Mamba module and the 2D Fourier Neural Operator (FNO2d) in the overall model to assess their contributions to reconstruction accuracy. Removing both FNO1d and FNO2d yields a MAE of 0.397 and a Max-AE of 0.355, indicating that despite Mamba's temporal modeling capability, the absence of spatial encoding significantly limits reconstruction performance. Introducing FNO1d alone leads to a substantial improvement, reducing MAE to 0.315 and Max-AE to 0.125. This demonstrates that FNO1d effectively captures transient variations by mapping temporal features to the frequency domain and integrating them via parameterized kernel operators. Notably, the reduction in Max-AE highlights FNO1d’s critical role in enhancing temporal-spatial interactions. Further adding FNO2d results in additional performance gains, with MAE and Max-AE dropping to 0.310 and 0.108, respectively. While the improvement is smaller, FNO2d contributes to more coherent global spatial structure and better local detail preservation. In summary, both FNO1d and FNO2d play complementary roles: FNO1d enhances temporal semantic fusion, while FNO2d refines spatial feature learning. Their combination significantly improves the reconstruction of complex, time-evolving physical fields.
\section{Conclusion}
In this paper, we introduced FR-Mamba, a novel approach for time-series physical field reconstruction using a State Space Model (SSM) integrated with Fourier Neural Operators (FNO). By efficiently capturing both temporal dependencies and spatial features, FR-Mamba significantly outperforms existing methods in reconstructing complex, transient physical fields such as fluid flow around a cylinder. Our method effectively addresses the limitations of traditional flow reconstruction techniques, particularly in terms of accuracy over long sequences and in regions with abrupt changes. The key contribution of this work lies in the design of the FNO-Mamba module, which combines the strength of SSM for long-term temporal modeling and FNO for global spatial feature extraction, providing a powerful tool for physical field reconstruction. Extensive experiments demonstrated that FR-Mamba achieves low Mean Absolute Error (MAE) and Maximum Absolute Error (Max-AE) across varying flow field conditions, establishing its robustness and versatility in dynamic flow environments. This approach is highly beneficial for applications in fluid dynamics, environmental monitoring, and other engineering domains where precise reconstruction of time-series physical fields is crucial. In future work, we plan to explore further optimizations and adapt the model to more complex physical phenomena, including multi-phase flows and higher-dimensional physical fields, with the goal of broadening its applicability to a wider range of scientific and engineering challenges.

\bibliographystyle{elsarticle-num} 
\bibliography{main}






\end{document}